\RequirePackage{fix-cm}
\documentclass[smallextended]{svjour3}       
\smartqed  

\usepackage{graphicx}
\usepackage{subfigure}
\usepackage{cite}
\usepackage{amsfonts}
\usepackage{multirow}
\usepackage{verbatim}
\usepackage{indentfirst}
\usepackage[shortlabels]{enumitem}
\setlist[enumerate]{nosep}
\usepackage{amsmath}
\begin{document}

\title{Transformer-Encoder-GRU (T-E-GRU) for Chinese Sentiment Analysis on Chinese Comment Text
}


\author{Binlong Zhang        \and
        Wei Zhou* 
}


\institute{Binlong Zhang \at
              Northwest University \\
              \email{binlong\_zhang@163.com}           
           \and
           *Corresponding Author: \\
           Wei Zhou \at
              Northwest University \\
              \email{mczhouwei12@gmail.com}
}

\date{Received: date / Accepted: date}

\maketitle

\begin{abstract}
Chinese sentiment analysis (CSA) has always been one of the  challenges in natural language processing due to its complexity and uncertainty.
Transformer has succeeded in capturing semantic features, but it uses position encoding to capture sequence features, which has great shortcomings compared with the recurrent model. In this paper, we propose T-E-GRU for Chinese sentiment analysis, which combine transformer encoder and GRU. We conducted experiments on three Chinese comment datasets. In view of the confusion of punctuation marks in Chinese comment texts, we selectively retain some punctuation marks with sentence segmentation ability. The experimental results show that T-E-GRU outperforms classic recurrent model and recurrent model with attention.
\keywords{Chinese reviews \and Chinese sentiment analysis \and Recurrent model \and Transformer-Encoder \and T-E-GRU}
\end{abstract}

\section{Introduction}
\label{intro}

Sentiment analysis has become one of the most active fields in natural language processing (NLP)~\cite{feldman2013techniques,zhang2018deep}. With the development of sentiment analysis, the research is not only limited to the computer field, but also expanded to management, social science, etc. 
Its potential commercial value has also attracted the attention of all sectors of society. With the development of GPU, sentiment analysis methods based on machine learning and deep learning gradually occupy the dominant position and continue to achieve the state-of-the-art performance. 

The earliest CSA is based on sentiment dictionary, which judges the sentiment tendency of the text through its vocabulary. But none of them have the ability to capture sequence features. Recurrent neural network (RNN) proposed by Jeffrey L Elman ~\cite{preRNN} is one of the effective methods to alleviate this defect, but it was accompanying by vanishing/exploding gradient problem~\cite{bengio1994Long-term-dependencies-diffcult}. Subsequently, a series of recurrent models based on RNN were proposed to address these problems, such as long short-term memory (LSTM)~\cite{hochreiter1997LSTM}, gate recurrent neural network (GRU)~\cite{cho2014learningGRU}, etc. However, the vanishing/exploding gradient problem on long sequence is still one of its most important limitations. To address the vanishing/exploding problems, Bahdanau et al.~\cite{bahdanau2014Attention-first} proposed the attention mechanism, which assigns weights to input sequence to make the model pay more attention to keywords. After that, the models with attention continues to achieve state-of-the-art performances. In particular, in 2017, Vaswani et al.~\cite{vaswani2017Transformer} proposed  scaled dot-product attention which has become one of the effective ways to apply attention mechanism and transformer model. Transformer and its variants have achieved the state-of-the-art performances on basic problems in many fields~\cite{BERT,vaswani2017Transformer}. Transformer-Encoder realizes the extraction of sequence features through position encoding (PE), which still has a gap with the natural sequence feature extractor such as RNN. 

Although sentiment analysis has made remarkable achievements in recent years, most of them are based on English. Compared with English, Chinese natural language processing is more challenging. On the one hand, Chinese has more complicated vocabulary and semantics. On the other hand, Chinese text semantics are more dependent on context. 
Inspired by the powerful global feature extraction ability of attention \& transformer and the powerful sequence feature extraction ability of recurrent model, we proposed Transformer-Encoder-GRU (T-E-GRU) which combine the GRU with the structure of transformer, and use it for sentiment analysis of Chinese reviews. We compare it with recurrent model, recurrent model with attention, etc. The results show that our model achieves better results.

The main contributions of this paper are summarized as following:
\begin{enumerate}
	\item We proposed a novel model called T-E-GRU based on GRU and transformer encoder, which is exactly suitable for CSA.
	\item Focus on the complexity and confusion of punctuation in Chinese comments, we have selectively retained some punctuation that has the ability to clauses.
	\item Extensive experiment demonstrate that T-E-GRU achieves better effect on CSA.
\end{enumerate}

\section{Related Work}

\subsection{Chinese Text To sequence}
In Chinese natural language processing, Chinese word segmentation (CWS) is a basic and important task, which segments text into independent word units. Compared with the natural delimiter spaces in English, Chinese has no formal delimiters and no strict division method, and even sometimes the division method depends on the context. The main disadvantages of dictionary-based word segmentation algorithms such as forward maximum matching (FMM), backward maximum matching (BMM), and bi-direction matching (BiMM) are slow in matching and cumbersome implementation of supplementing unrecorded words~\cite{CWS10yearReview}. Based on statistical machine learning algorithms, the main idea is that the more adjacent characters appear in the context at the same time, the more likely they are to form a word. The current main models are: N-gram model (N-gram)~\cite{N-gram}, hidden Markov model (HMM)~\cite{HMM}, condition random field (CRF)~\cite{CRF}, RNN variant model~\cite{LSTM-CWS,Bi-LSTM-CWS}. Various open CWS libraries such as Jieba, FoolNLTK, LTP, etc. have performed well in CWS. Although some people have raised doubts about the necessity of CWS in deep learning~\cite{WS-is-necessary}, most of the state-of-the-art models are based on CWS.

The simplest way to represent words is to use one-hot encoding which brings curse of dimensionality, semantic gap and poor scalability. Fighting the curse of dimensionality, Bengio Y et al. proposed a method of using low-dimensional vector to represent vocabulary~\cite{proposed-Embedding}. Word Embedding is a by-product of their model. In 2013, Tomas Mikolov proposed word2Vec~\cite{word2Vec}, which is a faster and better method for training word embedding model. There are two algorithms: skip-grams and continuous bag-of-words (CBOW)~\cite{CBOW}, both of which based on N-gram model.The N-gram model assumes that a word is only related to N surrounding words. This assumption determines that the disadvantage is insufficient use of global information. Jeffrey Pennington et al. proposed global vectors for Word representation (GloVe) which considers both global and local information~\cite{glove}. However, the word representation mentioned above is a static model, which cannot cope with the situation where a word has multiple meanings. In 2018, Peters M E et al. proposed embeddings from language model (ELMO)~\cite{ELMO} by using bidirectional LSTM. Since transformer has been proved to have powerful feature extraction ability in many researches~\cite{vaswani2017Transformer}, generative pre-training (GPT) was proposed~\cite{GPT}. It replaces LSTM in ELMO with transformer. Although GPT uses a unidirectional language model, it achieves the best results in 9 of the 12 tasks in NLP. In 2019, Devlin et al. proposed bidirectional encoder representations from transformers (BERT)~\cite{BERT} which replace unidirectional with bidirectional language models, and also combined with the tricks of CBOW. BERT, as synthesis of word representation model in recent years, achieved state-of-the-art performances in multiple NLP basic tasks.

\subsection{Recurrent models}
Recurrent model is one of the effective methods for processing sequence input. In 1990, neuroscientist Jeffrey L Elman proposed RNN which is based on the Jordan Network~\cite{preRNN} back-propagation (BP)~\cite{BP}. However, vanishing/exploding gradient problem on long sequence comes with RNN~\cite{bengio1994Long-term-dependencies-diffcult}. In 1997, Hochreiter et al. proposed LSTM, which can alleviate 
this problem by introducing delicate gate mechanism~\cite{hochreiter1997LSTM}. However, LSTM often has overfitting on some tasks due to the complexity of its structure. Cho et al. simplified the gate of LSTM and propose GRU~\cite{cho2014learningGRU}. GRU has the same performance as LSTM but with fewer parameters. RNN, LSTM and GRU have become the most important recurrent models, but none of them can completely solve the vanishing/exploding gradient problem. Up to now, methods to improve RNN from different perspectives are still proposed one after another~\cite{liu2021selfish}.

\subsection{Attention and Transformer}
In 2014, Mnih V et al. applied attention mechanism on RNN to solve image classification tasks~\cite{cv_Att}. Then, attention was used as a way to solve the information overload problem in NLP tasks. Bahdanau et al. extended the basic seq2seq model with attention, and it yields good results on longer sentences~\cite{bahdanau2014Attention-first}. Then Cheng proposed intra-attention~\cite{Att-breakout-Seq2seq}, which broke the limitation of attention mechanism used in seq2seq. Many researches~\cite{bahdanau2014Attention-first,Att-breakout-Seq2seq,cv_Att,RecurrentAttNet-AspectSentimentAnalysis} show that attention is effective in long sequence information loss. Vaswani et al. completely abandoned RNN and CNN to build transformer which is entirely based on the fully-connected layer and attention mechanism~\cite{vaswani2017Transformer}. In 2020, Zhou H et al. proposed probsparse self-attention mechanism and informer, and this model achieved high prediction capacity in the long sequence time-series forecasting~\cite{zhou2020informer}. Transformer is divided into two parts: encoding and decoding. Transformer-encoder uses position encoding to make the model understand sequence features. T-E-GRU replaces the position encoding with GRU, which has similar performance to LSTM. So T-E-GRU has the advantages of transformer-encoder and recurrent model, which will be more suitable for CSA.

\subsection{Chinese sentiment analysis}
Chinese sentiment analysis is very challenging due to the complexity of Chinese. A plain method is based on the n-gram model and syntax tree, using some classic machine learning classification models such as naive Bayesian (NB), support vector machine (SVM), etc. for CWS~\cite{zou2015sentiment}. The performance of their model were unsatisfactory. Ding et al.~\cite{ding2008holistic} also proposed a method of matching sentiment words in specific field for CSA. J Liang et al.~\cite{liang2015polarity} combined polarity shifting and LSTM for CSA. Xiao Z~\cite{xiao2016chinese} proposed bidirectional LSTM with word embedding for CSA. Their model already has high accuracy in CSA. As the attention mechanism and transformer were proposed, a series of models were  for CSA. Bin et al.~\cite{bin2017aspect} proposed multi-attention convolutional neural network (CNN) for aspect-based sentiment analysis. In 2019, Shi, Xiaoming et al.~\cite{shi2019attention} proposed attention-based bidirectional hierarchical LSTM networks. In 2021, Wang, Shitao et al.~\cite{wang2021bigru} proposed bigru-multi-head self-attention network for CSA. In addition, many state-of-the-art models based on transformer or BERT have been proposed for CSA~\cite{tang2021chinese,zhu2019transformer,gao2021chinese}.

\section{T-E-GRU}
At present, the state-of-the-art models for NLP are mainly based on transformer or recurrent model. In general, transformer has powerful global feature extraction capabilities and is very suitable for CSA. Recurrent model is a natural sequence structure, which is very suitable for capturing sequence features. In particular, GRU has become one of the best recurrent models due to its similar performance to LSTM and cheaper calculations.

Inspired by the global feature extraction capabilities of transformer and the sequence feature extraction capabilities of GRU, we proposed T-E-GRU for CSA. T-E-GRU combines transformer-encoder and GRU. Compared with recurrent model, the multi-head self-attention mechanism and residual connection in the transformer-encoder of T-E-GRU can better cooperate with the loss of long sequence information. Different from transformer, T-E-GRU uses GRU to extract sequence features instead of position encoding, which can better deal with the problem that text sentiment depends on word order. In addition, compared to recurrent model with attention, T-E-GRU combines the transformer structure that has achieved good results in NLP.

\begin{figure}
	\centering
	\includegraphics[width=1\textwidth]{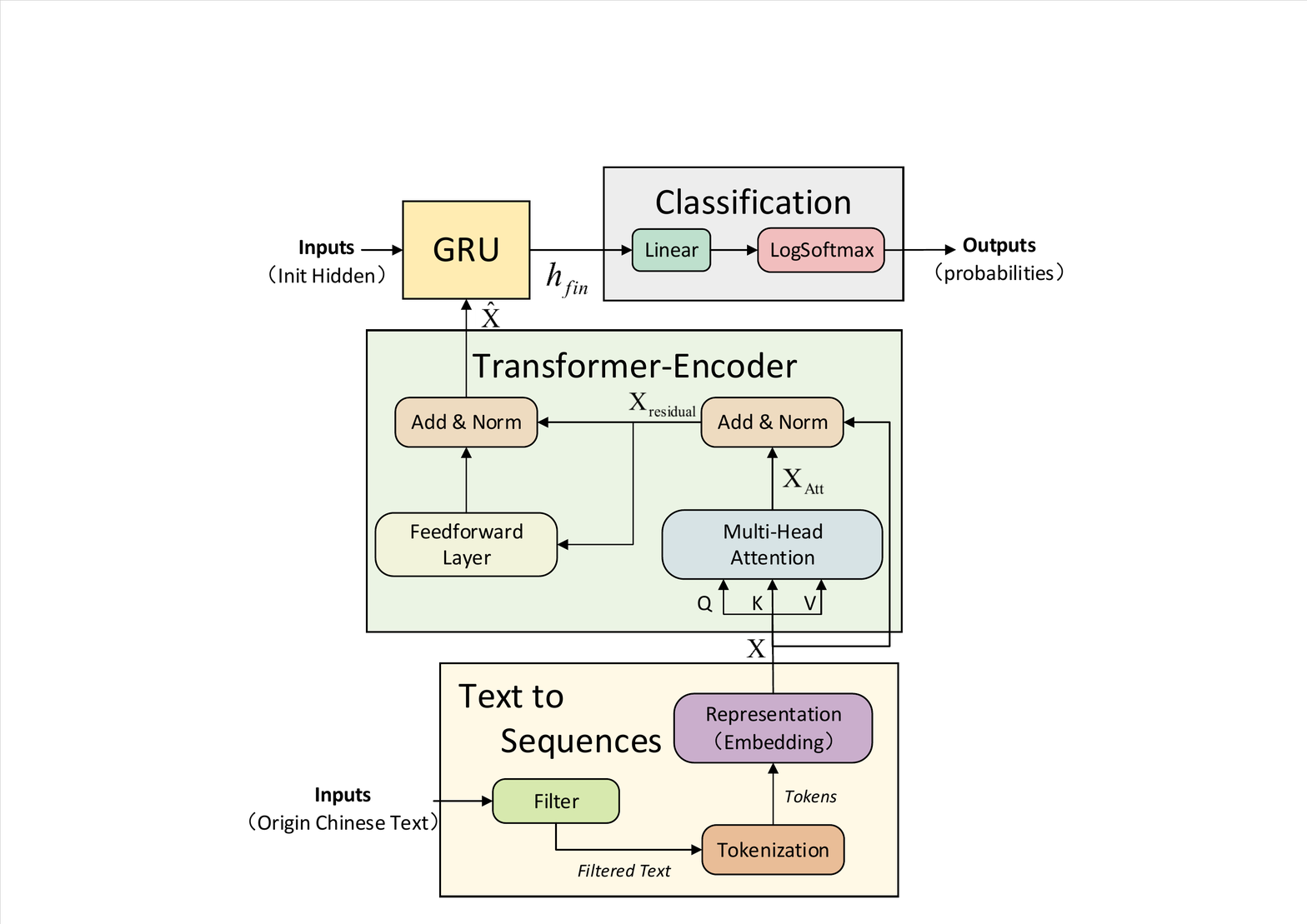}
	\caption{Transformer-Encoder-GRU}
	\label{fig:Transformer-Encoder-GRU}       
\end{figure}

As shown in Figure~\ref{fig:Transformer-Encoder-GRU}, our model consists of four parts: text to sequence, transformer-encoder, GRU and classification. Text to sequence converts original Chinese text into vector sequence. Transformer-encoder reprocesses the original vector sequence So that each processed vector is determined by the entire input vector sequence. GRU uses the characteristics of recurrent model to process sequence features. Due to the reprocessing in the transformer-encoder, processed vector has global information, which can better deal with the problem of long sequence information loss, and this problem is the most difficult one to overcome in recurrent model. Finally, the final state of the GRU is adopted as the input of the classification to output the predicted probability.

\subsection{Text To Sequence}
Text to sequence maps the original Chinese text to vector sequence $X = (x_1,x_2,\cdots,x_n)$, where $x_i\in \mathbb{R}^{d_{R}} $. n is the number of tokens and $d_R$ is the dimension of word representation. Text to sequence mainly consists of three parts: filter, tokenization, representation.
Filter is used to deal with irregular punctuation, characters, etc. in the original Chinese text. After filtering, the tokens generated after tokenization will be shorter than before and will be fewer meaningless tokens. Specifically, the filter selectively removes most of the punctuation that does not have the ability to clause. Then tokenization convert the filtered text into a series of tokens. Here, any suitable CWS library can be used, and we use Jieba to finish CWS. Next, we use vectors to represent these tokens. Word embedding is one of the important representation methods. Representation takes these tokens as input and vector sequence $X = (x_1,x_2,\cdots,x_n)$ as output. After the above processes, text to sequence converts the original Chinese text into vector sequence $X$.

\subsection{Transformer-Encoder}
Transformer-encoder takes $X$ from text to sequence as input and generates output $\hat{X} = (\hat{x}_1,\hat{x}_2,\cdots ,\hat{x}_n)$. As shown in Figure~\ref{fig:Transformer-Encoder-GRU}, transform-encoder is mainly divided into multi-head self-attention and feedforward network.

Attention mechanism has become one of the most effective methods to capture global information~\cite{attentionInMulTask,attentionInNLP,vaswani2017Transformer}. Multi-head attention is based on scaled dot-product attention proposed by Vaswani et al.~\cite{vaswani2017Transformer}. Multi-head attention is shown in the following equations:
\begin{equation}
	\begin{split}
		&MultiHead(Q,K,V)=concat(Att_1,Att_2,...,Att_n)\\
		&where \ \ Att_i =softmax(\frac{QK^T}{\sqrt{d_k}})V
	\end{split}
\label{attention}
\end{equation}
where $Q,K,V$ represent query, keys and values respectively which they are all input matrices, $d_k$ represents the dimension of the keys, $n$ equals to the number of heads, and we set $n=2$ in our model. Especially when $Q,K,V$ are the same, it is called self-attention. Here we use the output $X$ from text to sequence as $Q,K,V$, and then output $X_{Att}$. Then we use residual connection and feedforward network  in the following:
\begin{equation}
	\begin{split}
	    &X_{Att} = MultiHead(X,X,X)\\
		&X_{residual} = norm(X_{Att} + X)\\
		&\hat{X} = norm(X_{residual}+\text{FFN}(X_{residual}))
	\end{split}
\label{equation:T-E}
\end{equation}
where $norm$ is normalization layer, $\text{FFN}$ consists of two linear transformations with a ReLU:
\begin{equation}
	\text{FFN}(X_{residual}) = Linear(\max(0,Linear(X_{residual})))
\end{equation}
where the dimension of inner-layer is 2048 and other details of transformer-encoder are mentioned in~\cite{vaswani2017Transformer}.

Finally, transformer-encoder transfers $X = (x_1,x_2,\cdots,x_n)$ to  $\hat{X} = (\hat{x}_1,\hat{x}_2,\cdots ,\hat{x}_n)$, where $\hat{x_i}$ is determined by the entire input $X$.

\subsection{GRU}
\begin{figure}
	\centering
	\includegraphics[width=1\textwidth]{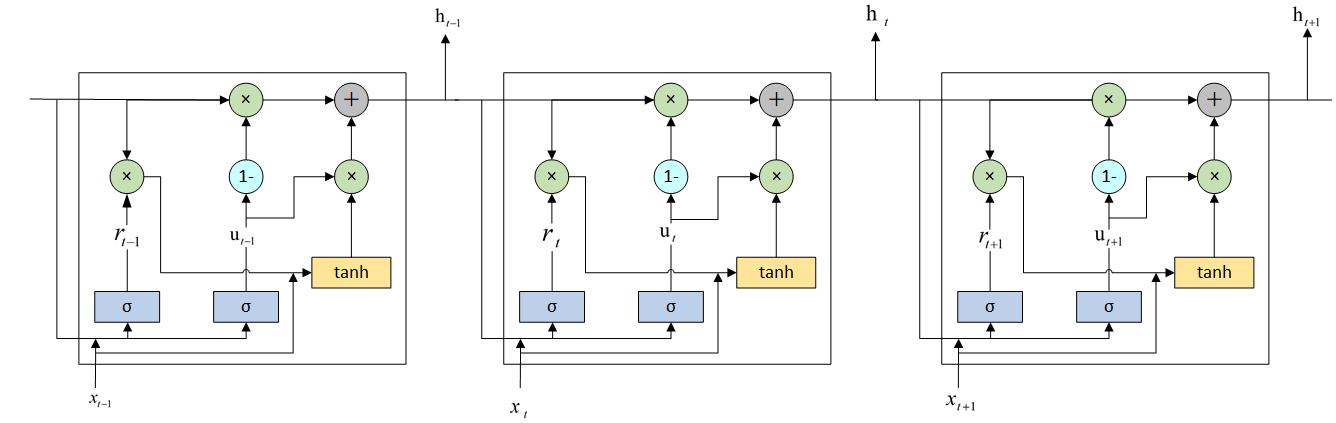}
	\caption{GRU}
	\label{fig:GRU}       
\end{figure}
GRU is one of the variants of RNN, which improved from the LSTM. It performs similarly to LSTM, but is computationally cheaper. As shown in Figure~\ref{fig:GRU}. For any input state $x_t$ and the hidden state $h_{t-1}$  generated by the previous sequence:
\begin{equation}
	\begin{split}
		&u_t = sigmoid(x_tW_z+h_{t-1}U_z) \\
		&r_t = sigmoid(x_tW_r+h_{t-1}U_r) \\
		&h_t = (1-u_t)h_{t-1} + u_t*tanh(x_tW_h+(h_{t-1} \cdot r_t)U_h)
	\end{split}
\end{equation}
where $r_t$ is the reset gate, $u_t$ is the update gate,$W_{z,r,h}$  and $U_{z,r,h}$ are the weight matrices that need to be trained. 

Here we take $\hat{X}$ from transformer-encoder as input of GRU to generate the state sequence $(h_1,h_2,\cdots,h_n)$. We use the final state $h_n$ as the output of GRU, and then use the classification to implement CSA. The classification here is simply composed of a linear layer and logsoftmax.

\section{Experimental}
In this section, we compare our method with the state-of-the-art frameworks (Including various recurrent neural networks and recurrent neural networks with attention) on three datasets for CSA tasks, including DMCS\_V2~\ref{sec:dmcs}, YF\_DianPing~\ref{sec:yf_dianping} and Online\_shopping\_10 cats~\ref{sec:online_shopping} datasets. We also conduct ablation study to research the performances of T-E-GRU.

\subsection{DMCS\_V2}
\label{sec:dmcs}
\textbf{Dataset.} Douban movie short comments dataset V2 (DMSC\_V2\footnote{https://www.kaggle.com/utmhikari/doubanmovieshortcomments}) comes from the Kaggle competition. It collects user reviews, ratings, and likes of movies on Douban (a platform for Chinese film lovers). 
This dataset collected short user reviews of 28 popular movies, with a total of more than 2 million comments.
We label 1, 2-star reviews as “negative reviews” and 4,5-star reviews as “positive reviews” (3-star reviews are ignored because of its vague sentimentality). In the experiments we randomly selected 700,000 comments data including 350,000 positive samples and 350,000 negative samples. We divide them into three parts, 480,000 for the training set, 120,000 for the validation set, and 100,000 for the test set.

\textbf{Data preprocessing.}
\label{data-preprocessing-DMCS}
Firstly, we need to deal with the punctuation in the text data. Since the use of punctuation marks in most short film reviews is not standardized, and some punctuation has less semantic information, we retain some punctuation with clause ability, such as comma, full stops, question marks, semicolons, etc. and deleted others. Our focus is not to discuss CWS model or the necessity of CWS in deep learning models, so we use open source library Jieba developed by Baidu engineer Sun Junyi. After word segmentation, we pad and truncate the tokens for batch processing. In addition, we use pre-trained word embedding to convert words into vectors, which was trained by Li S, Zhao Z, Hu R, et al.~\cite{preTrainEmbedding} by using skip-gram with negative sample method based on Zhihu Q\&A~\cite{preTrainEmbedding}. It contains 260,000 used high-frequency words, which can cover 93.98\% of the words in train set. The dimension of the word vector is 300. DMCS\_V2 is a short movies comments dataset. After tokenization, 99.8\% of samples' tokens are less than 100. So we align the input sequence to 100 by padding and truncating it at the front of it. For word representation, we use 200,000 high-frequency words in the pre-trained word embedding model, which can cover 93.86\% of the word needed in there. 

\textbf{Train Details.}
\label{Train-hyper-parameters}
In the experiments of RNN, LSTM and GRU, we use negative log-likelihood (NLL) loss function, set the batch size and the hidden size to 128 and 256 respectively, adopt Stochastic Gradient Descent (SGD) optimizer with learning rate of 0.002 for 100 epochs, and decay the learning rate by 0.5 every 50 epochs.
We use the same settings for the bi-directional recurrent model, and the model combined with attention mechanism and recurrent models as mentioned in before, except that the hidden layer in a single direction is set to 128, and set the learning rate to 0.01. Finally, we also use the same setting on our T-E-GRU, and set the dropout of transformer-encoder to 0.3.

\textbf{Results and Analysis.}
\label{sec:Douban_result}
\begin{table}
	\caption{The performance of various models on the DMCS\_V2}
	\label{tab:DMSC_v2-Result}       
	\begin{tabular}{llll}
		\hline\noalign{\smallskip}
		Method 		                &Accuracy 		    &F1			        &Test Time(ms)\\
		\noalign{\smallskip}\hline\noalign{\smallskip}
	    RNN~\cite{RNN-first}	        &88.80\%	        &88.77\%            &0.8483\\
		LSTM~\cite{hochreiter1997LSTM}  &88.41\% 			&88.28\% 	        &2.0053\\
		GRU~\cite{cho2014learningGRU}	&88.70\%  			&88.80\%	        &1.9884\\
		Bi-RNN						    &88.30\%			&88.30\%	        &1.4551\\
		Bi-LSTM						    &87.95\%			&88.02\%	        &3.8016\\
		Bi-GRU						    &88.00\%			&88.80\%	        &1.4989\\
		\noalign{\smallskip}\hline\noalign{\smallskip}
		RNN-Attention				    &88.50\%			&88.34\%	        &1.2503\\
		LSTM-Attention	                &89.43\%	        &89.32\%	        &2.3592\\
		GRU-Attention				    &88.08\%	        &88.27\%	        &2.3599\\
		Attention-RNN				    &85.77\%			&86.09\%	        &1.2624\\
		Attention-LSTM				    &85.67\%			&85.55\%	        &2.4111\\
		Attention-GRU				    &84.68\%			&84.32\%	        &2.3835\\
		BiRNN-Attention			        &88.64\%			&88.50\%	        &1.8544\\
		BiLSTM-Attention			    &88.45\%			&88.48\%	        &4.3638\\
		BiGRU-Attention			        &88.18\%			&87.60\%	        &1.9140\\
		\noalign{\smallskip}\hline\noalign{\smallskip}
		\textbf{T-E-GRU}		        &90.09\%	        &90.07\%	        &2.6987\\
		\noalign{\smallskip}\hline
	\end{tabular}
\end{table}
The performance of the above models on the test set is shown in Table~\ref{tab:DMSC_v2-Result}. 
We use accuracy and F1 as the criteria: 
\begin{equation}
    \label{equation:Acc&F1}
	\begin{split}
	&Acc = \frac{TP+TN}{TP+TN+FP+FN}\\
	&F1 = \frac{2 \times TP}{2\times TP +FP +FN}
	\end{split}
\end{equation}
where TP, TN, FP and FN represent true positive, true negative, false positive, and false negative respectively. Test Time denotes the time required for trained model to test a comment.
As shown in Table~\ref{tab:DMSC_v2-Result}, the accuracy and F1 of the Bi-directional recurrent model are 0.1\%$\sim$1\% lower than recurrent model on DMSC\_V2. The best recurrent model is simple RNN, and its accuracy and F1 have reached 88.8\% and 89.0\% respectively. At the same time, simple RNN also has a good performance in time with 0.8483ms.
After introducing attention, the accuracy, F1 of various recurrent model have not improved much, and it also slow down the testing of the model. In addition, the improvement is only effective when the attention layer is after the recurrent layer, and the others have a slight decrease. The best recurrent model with attention is LSTM-Attention, its accuracy and F1 are 89.43\% and 89.32\%, which is about 0.5\% improvement compared to the best recurrent model. 
However, compared with LSTM-Attention, T-E-GRU has an improvement of about 0.1\%. In other word, T-E-GRU is the best in experimental model, with 90.09\% accuracy and 90.07\% F1. The drawback is that T-E—GRU costs 6\% more test time than LSTM-Attention.

In general, we can find on DMSC\_V2:
\begin{enumerate}
    \item The simple RNN got the best performance in recurrent models.
    \item LSTM-Attention is the best when recurrent model is combined with attention,but it also leads to more time cost.
    \item T-E-GRU achieved the best performance in the models of mentioned above, with 90.09\% accuracy and 90.07\% F1. And the time required for model testing is very similar to LSTM-Attention.
\end{enumerate}


\subsection{YF\_DianPing Dataset}
\label{sec:yf_dianping}

\textbf{Dataset.}
YF\_DianPing~\cite{DianpingData} is customer reviews of restaurant, which has more characters and more complicated than DMSC\_V2.
It contains the user reviews crawled from a famous Chinese online review webset DianPing.com, including the 3,605,300 reviews of 510,071 users. After eliminating the NaN data, we randomly selected 80000 1-points reviews as positive samples, and 80,000 1-points reviews as negative samples.  We divide  them  into  two  parts, 80\% are used as the training set and 20\% are used as the validation set. Then we obtained the test set by taking 30,000 4-points reviews as positive samples and 30,000 2-points reviews with as negative samples. 

\textbf{Data preprocessing.}
The processing of punctuation, CWS, the padding/ truncating and pre-trained word embedding model are the same as Section~\ref{data-preprocessing-DMCS}.  After tokenization, the longest sample has 1386 tokens and 96.2\% of the samples' tokens are less than 400. So we align the input sequence to 400 by using the same method as in Section~\ref{data-preprocessing-DMCS}. For word representation, we use 150,000 high-frequency words, which covering 86.0\% of the words in the training set. Due to the complexity of the words used in the data set, using of 260,000 high-frequency words only covers 86.5\%, so it is more appropriate to use 150,000 high-frequency.

\textbf{Train Details.}
On YF\_DianPing Dataset, we did the same experiment and only modified the dataset. The various hyperparameters, loss functions, optimizers, etc. of each model are the same as Section~\ref{Train-hyper-parameters}.

\textbf{Results and Analysis}
\begin{table}
	\caption{The performance of various models on the YF\_DianPing}
	\label{tab:DianPing-Result}       
	\begin{tabular}{llll}
		\hline\noalign{\smallskip}
	    Method 						& Accuracy 		    &F1			        &Test Time(ms)\\
		\noalign{\smallskip}\hline\noalign{\smallskip}
	    RNN~\cite{RNN-first}    	    &87.95\%            &88.38\%            &3.0484\\
		LSTM~\cite{hochreiter1997LSTM}	&88.40\% 			&88.91\% 	        &7.4615\\
		GRU~\cite{cho2014learningGRU}	&88.83\%  			&89.44\%	        &7.5326\\
		Bi-RNN						    &89.23\%			&89.51\%	        &5.5511\\
		Bi-LSTM						    &87.90\%			&87.50\%	        &14.6120\\
		Bi-GRU						    &89.55\%	        &89.94\%	        &5.3284\\
		\noalign{\smallskip}\hline\noalign{\smallskip}
		RNN-Attention				    &88.92\%			&89.46\%	        &3.6268\\
		LSTM-Attention				    &90.45\%            &90.81\%	        &7.8170\\
		GRU-Attention				    &88.55\%			&89.07\%	        &7.7112\\
		Attention-RNN				    &87.29\%			&88.12\%	        &3.2191\\
		Attention-LSTM				    &87.92\%			&88.54\%	        &8.2530\\		
		Attention-GRU				    &88.06\%			&88.64\%	        &8.1179\\
	    BiRNN-Attention			        &89.54\%			&89.86\%	        &5.6455\\
		BiLSTM-Attention			    &87.70\%			&88.20\%	        &15.5177\\
		BiGRU-Attention			        &88.88\%			&89.36\%	        &8.6010\\
		\noalign{\smallskip}\hline\noalign{\smallskip}
		\textbf{T-E-GRU}			    &90.76\%	        &90.98\%	        &8.0009\\	
		\noalign{\smallskip}\hline
	\end{tabular}
\end{table}
The results of various models on the DianPing test set are shown in Table~\ref{tab:DianPing-Result}. The test set uses 2 points and 4 points review data, while the training set uses 1 point and 5 points review data, so there may be more ambiguous texts in the test set, resulting in poor performances. 
On this test set, Bi-GRU has the best performance in recurrent model, with accuracy and F1 reaching 89.55\% and 89.94\% respectively. Compared with the best-performing model RNN in Douban data, Bi-GRU has an improvement of about 3\% on both metrics. 
LSTM-Attention achieves the best performances in attention-based recurrent model with 90.45\% accuracy and 90.81\% F1.
T-E-GRU is slightly better than LSTM-Attention in both F1 and accuracy. Although not much improvement, T-E-GRU still achieved the best performance among all tested models with 90.76\% accuracy and 90.98\% F1. Similar to the results on Doublan, T-E-GRU still takes about 2\% more test time than LSTM-Attention on this dataset.

In general, We can find:
\begin{enumerate}
    \item RNN had a good performance on Douban before, in YF\_DianPing it did not achieve good performance due to the longer tokens. The Bi-GRU has achieved the best performance in recurrent model.
    \item LSTM-Attention is still the best performance in recurrent model with attention.
    \item Although not much improvement, T-E-GRU still achieved the best performance among all tested models with 90.76\% accuracy and 90.98\% F1. This also shows that T-E-GRU can still perform well compared to other models in the face of a more complex test set than training set. And its test time is close to LSTM-Attention.
\end{enumerate}

\subsection{Online\_shopping\_10 cats Dataset}
\label{sec:online_shopping}

\textbf{Dataset.}
This dataset comes from various Chinese E-commerce platforms, with a total of more than 60,000 comment data, including 30,000 positive and negative comments. We ignored their category tags and only used sentimental tags. We divide them into training set, validation set, and test set according to the ratio of 6:1:3. This dataset is smaller than the previous two datasets, and we use it to test the performance of our model on small-scale datasets.

\textbf{Data preprocessing.}
Compared with the previous two datasets, it is a small-scale moderate-length dataset. After tokenization, the longest sample has 1502 tokens and 99.0\% of the samples' tokens are less than 200. So we align the input sequence to 200. For word representation, we use 200,000 high-frequency words, which can cover 94.9\% of the word needed in the training set. Using all words in pre-trained word embedding model is only about 0.2\% improvement. 

\textbf{Train Details.}
In the experiments of Online\_shopping\_10 cats dataset, the various hyperparameters, loss functions, optimizers, etc. of each model are the same as Section~\ref{Train-hyper-parameters}. However, since the size of this data set is small, many models use 100 epochs can not complete the training. To solve this, we double the initial learning rate and set the training epochs to 300. The performances of each model are shown in Table~\ref{tab:OnlineShopping-Result}, where 300 epochs of training are marked after the model name.

\textbf{Results and Analysis.}
\begin{table}
	\caption{The performance of various models on the Online\_Shopping}
	\label{tab:OnlineShopping-Result}       
	\begin{tabular}{lllllll}
		\hline\noalign{\smallskip}
		Method 						& Accuracy 	    	&F1		            &Test Time(ms)\\
		\noalign{\smallskip}\hline\noalign{\smallskip}
	    RNN	 						&89.68\%            &89.93\%            &1.5092\\
		LSTM						&90.21\%            &90.06\%            &3.7045\\
		GRU	 						&90.08\%            &90.36\%            &3.7701\\
		Bi-RNN					    &90.10\%            &90.07\%            &2.6567\\
		Bi-LSTM						&90.10\%            &90.10\%            &7.1392\\
		Bi-GRU						&90.34\%            &90.13\%            &2.8293\\
		\noalign{\smallskip}\hline\noalign{\smallskip}
		RNN-Attention				&90.07\%			&90.10\%	        &1.9176\\
		LSTM-Attention				&91.86\%	        &91.76\%	        &4.0548\\
		GRU-Attention-300			&90.93\%			&90.96\%	        &3.9876\\
		Attention-RNN				&90.68\%			&90.72\%	        &1.8351\\
		Attention-LSTM-300			&90.91\%			&90.91\%	        &4.1661\\
		Attention-GRU-300			&89.56\%			&89.65\%	        &4.1363\\
	    BiRNN-Attention-300		    &92.32\%	        &92.38\%	        &3.0296\\
		BiLSTM-Attention-300		&90.28\%			&90.16\%	        &7.7145\\
		BiGRU-Attention-300			&91.09\%			&91.05\%	        &3.2287\\
		\noalign{\smallskip}\hline\noalign{\smallskip}
		\textbf{T-E-GRU}		    &92.92\%	        &93.05\%	        &4.3890\\
		\noalign{\smallskip}\hline
	\end{tabular}
\end{table}
The results of various models on the Online\_shopping\_10 cats test set are shown in Table~\ref{tab:OnlineShopping-Result}. On this test set, various recurrent models have similar performance. The accuracy and F1 of recurrent models are around 90\% and 90\%. BiRNN-Attention has the best performance in recurrent model with attention, with accuracy and F1 reaching 92.32\% and 92.38\% respectively. Compared with the recurrent model, BiRNN-Attention has an improvement of about 2\%. In contrast, LSTM-Attenion, which performed well before, was defeated on this data set. Compared with BiRNN-Attention, T-E-GRU has about 0.6\% and 1\% improvement in accuracy and F1 respectively, and it takes 40\% more in test time.

In general, we can find:
\begin{enumerate}
    \item On this data set, Bi-GRU still achieved the best performance in recurrent model with 90.34\% accuracy and 90.13\% F1. But compared to other recurrent models, the improvement of Bi-GRU is ambiguous.
    \item BiRNN-Attention has achieved the best results in recurrent model with attention. Compared with recurrent model, BiRNN-Attention has about 2\% improvement in accuracy and F1.
    \item T-E-GRU still gains the best accuracy and F1, which are 92.92\% and 93.05\% respectively. And it only takes 40\% more time than BiGRU-Attention.
\end{enumerate}

\subsection{Ablation Study}

We conducted ablation experiments to better understand the contributions and the hyperparameter setting of T-E-GRU. We trained the network and tested it on the DMSC\_V2, YF\_DianPing and Online\_shopping datasets.

\subsubsection{Replacing GRU with other recurrent layer}

\begin{table}
	\caption{The performance of replacing GRU with other recurrent layers}
	\label{tab:T-E-OtherRNN}       
	\begin{tabular}{llllllll}
		\hline\noalign{\smallskip}
		Dataset    &Method 		& Accuracy 		&F1		 & Test Time(ms)\\
		\noalign{\smallskip}\hline\noalign{\smallskip}
		\multirow{3}{*}{DMSC\_V2}
		&T-E-RNN			&89.25\%			&89.30\%	      &1.5390\\
		&T-E-LSTM			&89.75\%			&89.88\%	      &2.9321\\
		&T-E-BiRNN          &89.26\%            &89.30\%          &2.1434\\
		&T-E-BiLSTM			&89.81\%			&89.74\%	      &4.7107\\
		&T-E-BiGRU			&89.77\%			&89.85\%	      &2.2185\\
		&\textbf{T-E-GRU}	&90.09\%	        &90.07\%	      &2.6987\\

		\noalign{\smallskip}\hline\noalign{\smallskip}
		\multirow{3}{*}{YF\_DianPing}
		&T-E-RNN			&88.91\%			&89.42\%	    &3.4685\\
		&T-E-LSTM			&90.25\%			&90.70\%	    &8.0919\\
		&T-E-BiRNN          &88.94\%            &89.40\%        &5.9280\\
		&T-E-BiLSTM			&90.63\%			&90.92\%	    &15.4538\\
		&T-E-BiGRU			&90.73\%			&90.89\%	    &6.3677\\
		&\textbf{T-E-GRU}	&90.76\%	        &90.98\%	    &8.0009\\
		\noalign{\smallskip}\hline\noalign{\smallskip}
		\multirow{3}{*}{Online\_shopping}
		&T-E-RNN			&91.16\%	        &91.01\%	    &2.1586\\	
		&T-E-LSTM			&92.43\%	        &92.59\%	    &4.4119\\	
		&T-E-BiRNN          &91.40\%            &91.49\%        &3.3178\\
		&T-E-BiLSTM			&92.41\%			&92.47\%	    &8.2653\\
		&T-E-BiGRU			&92.38\%			&92.60\%	    &3.3085\\
		&\textbf{T-E-GRU-300}&92.92\%	        &93.05\%	    &4.3890\\
		\noalign{\smallskip}\hline\noalign{\smallskip}
	\end{tabular}
\end{table}
In this section, we tried replacing GRU with other recurrent models (RNN, LSTM). The experimental results are shown in Table~\ref{tab:T-E-OtherRNN}.
On DMSC\_V2, T-E-GRU has achieved the best performance compared to other models, but it has only a slight improvement in accuracy and F1. 
On YF\_DianPing, 
T-E-GRU only has less than 1\% improvement in accuracy, F1 than other models. 
T-E-GRU is still the best accuracy and F1, which are 90.76\% and 90.98\% respectively. 
On Online\_shopping, T-E-GRU still obtain the best accuracy and F1,
In general, GRU is more suitable for combining with transformer-encoder.The multiple of the time spent is usually positively correlated with the number of tokens.

\subsubsection{Adjustment of hyperparameters in T-E-GRU}

\begin{figure}
	\centering
	\subfigure[DMSC\_V2.]{
	\includegraphics[width=0.45\textwidth]{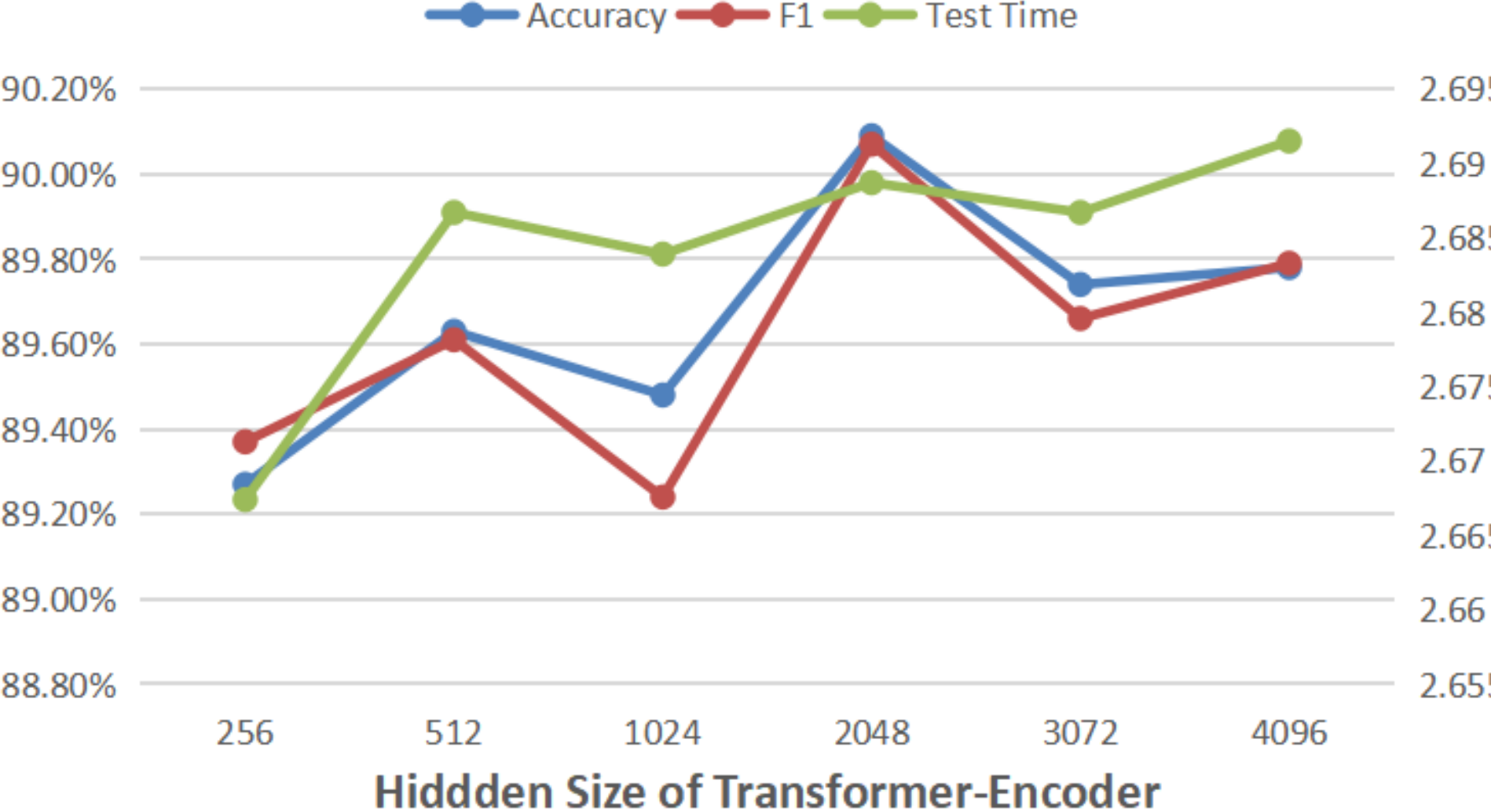}

    }
    \quad
	\subfigure[YF\_DianPing]{
	\includegraphics[width=0.45\textwidth]{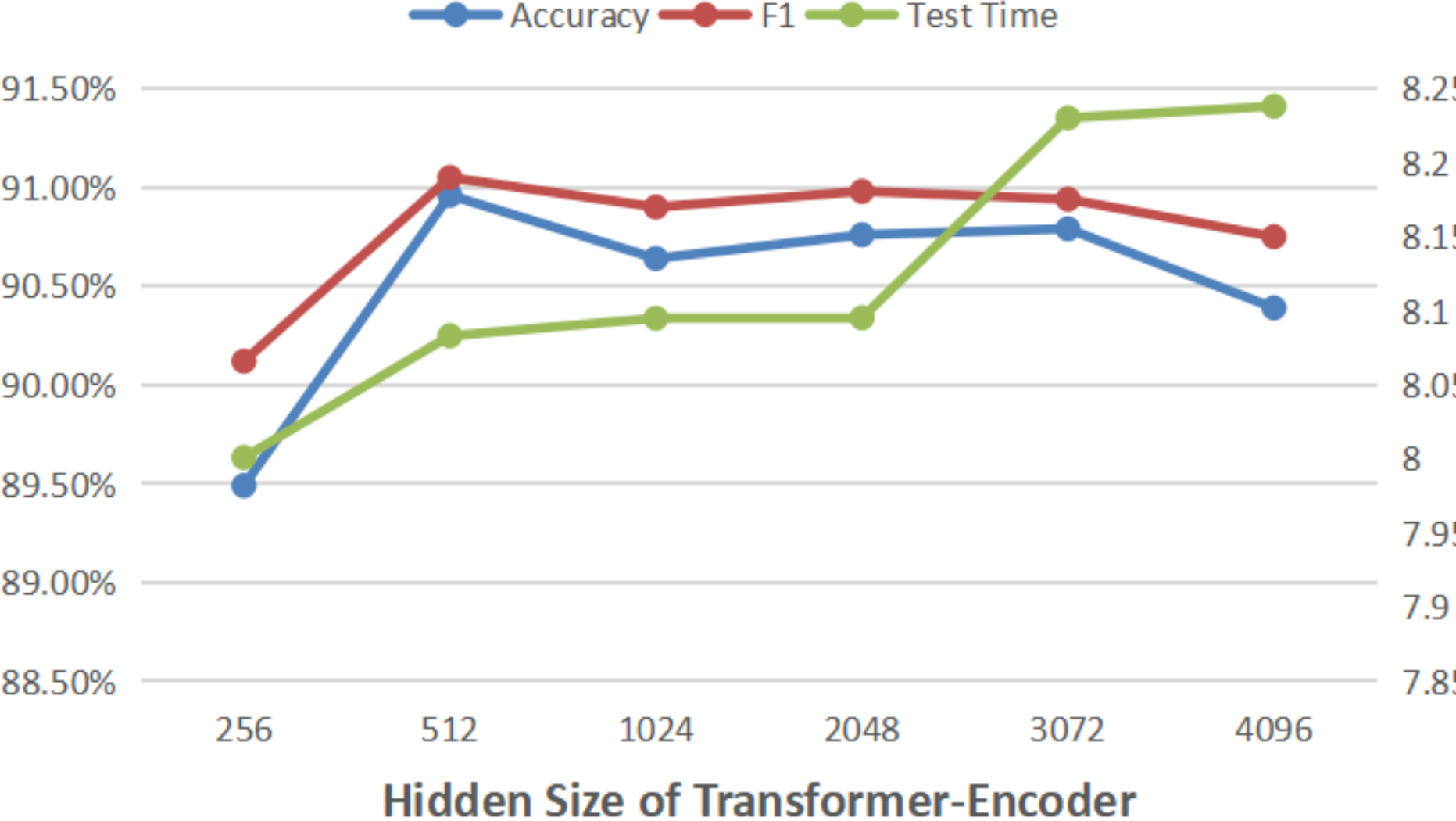}
    }
    \quad
	\subfigure[Online\_shopping\_10 cats Dataset]{
	\includegraphics[width=0.45\textwidth]{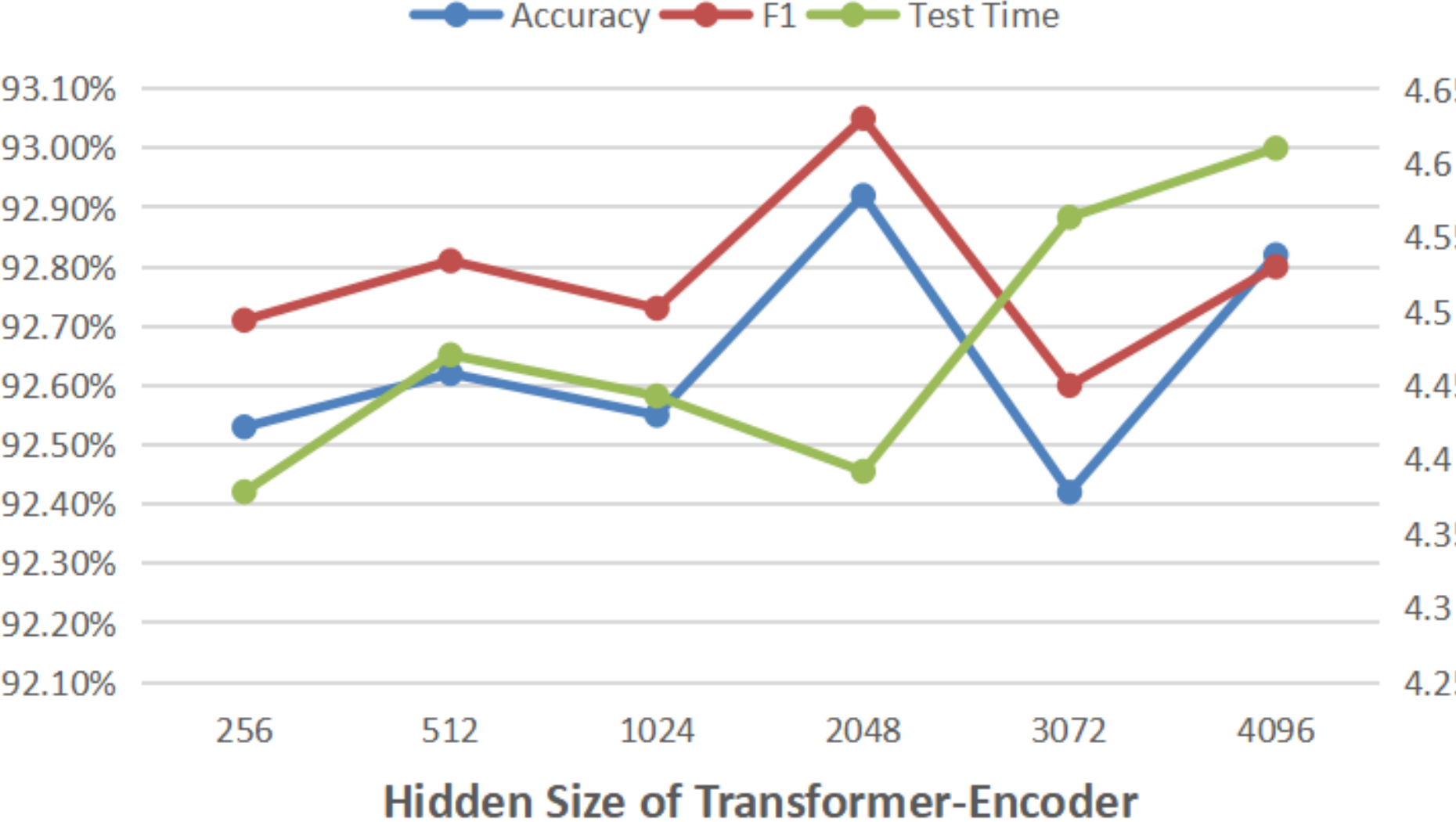}
    }
	\caption{Hidden size of Transformer-Encoder}
	\label{fig:Hidden_size_Transformer-Encoder}       
\end{figure}
Firstly, we tried to adjust the hidden size of the transformer-encoder. The results on three datasets mentioned in Section~\ref{sec:dmcs},~\ref{sec:yf_dianping},~\ref{sec:online_shopping} are shown in Figure~\ref{fig:Hidden_size_Transformer-Encoder}. The accuracy, F1, and Test Time are introduced in Section~\ref{sec:Douban_result}. From the results, we find that the different hidden sizes of transformer-encoder have little effect on Test Time, mostly less than 0.5(ms). For accuracy and F1, DMSC\_V2 and Online\_shopping have the best performance in 2048, while 2048 and 3072 on YF\_DianPing have very similar performance.

\begin{figure}
	\centering
	\subfigure[DMSC\_V2.]{
	\includegraphics[width=0.45\textwidth]{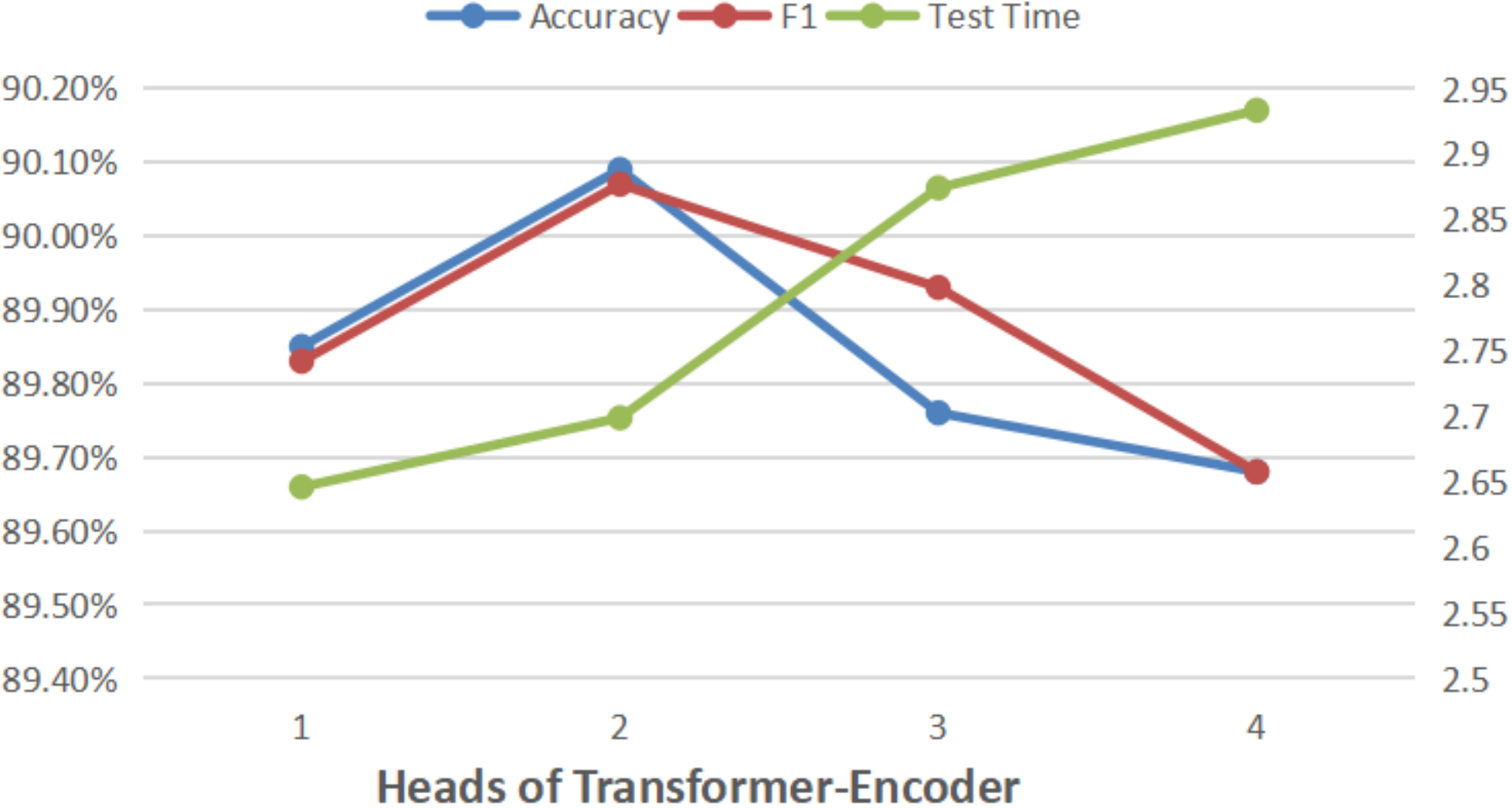}

    }
    \quad
	\subfigure[YF\_DianPing]{
	\includegraphics[width=0.45\textwidth]{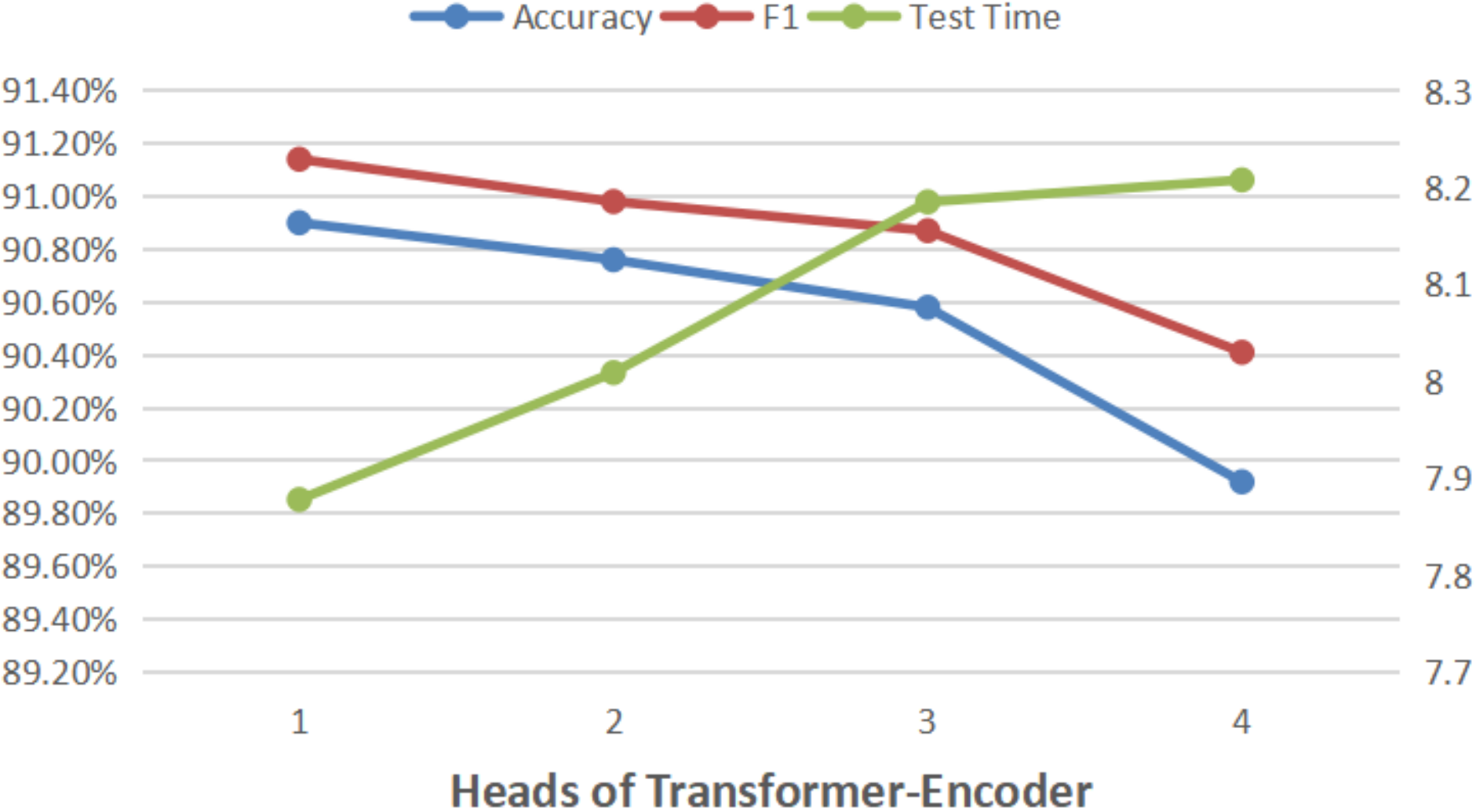}
    }
    \quad
	\subfigure[Online\_shopping\_10 cats Dataset]{
	\includegraphics[width=0.45\textwidth]{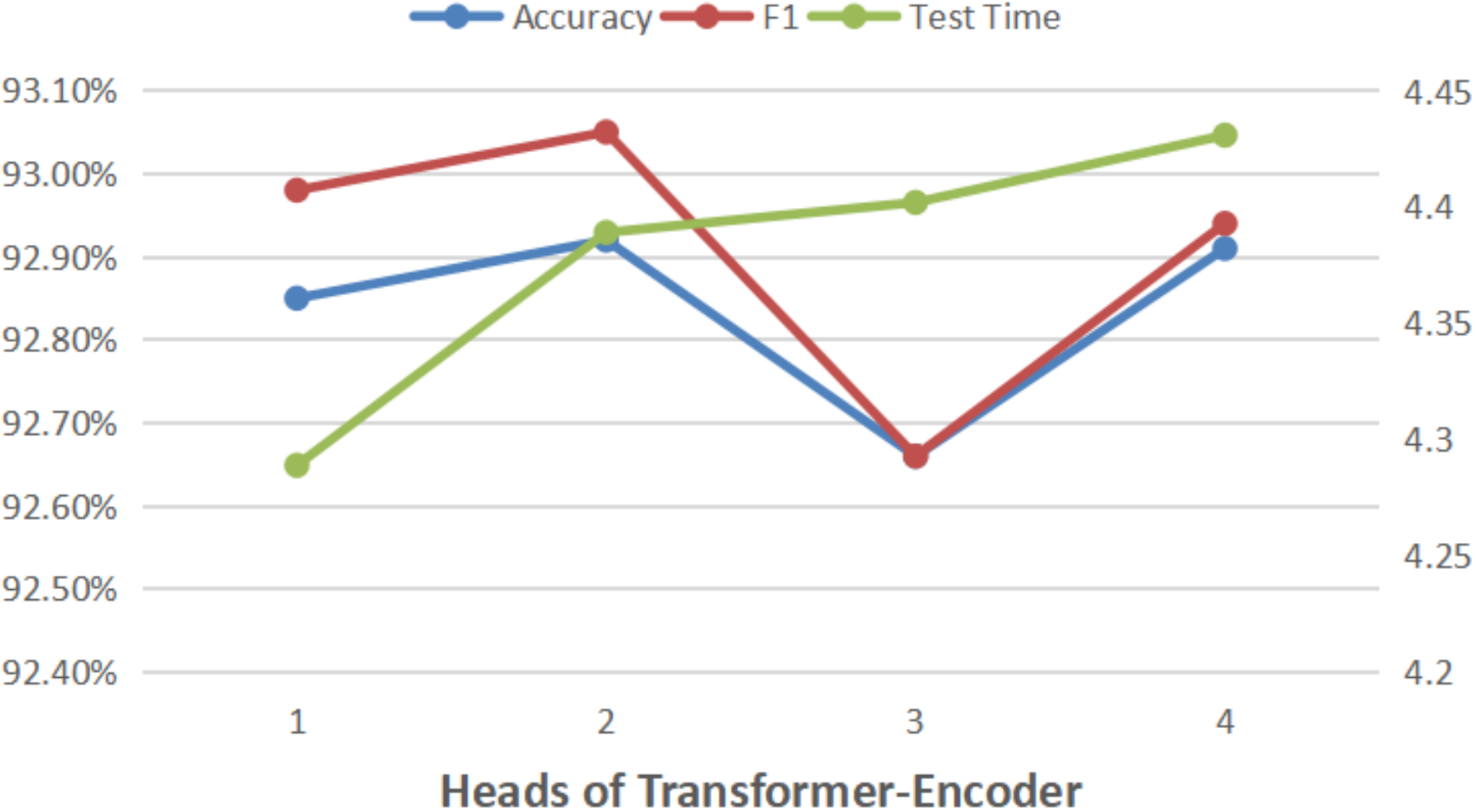}
    }
	\caption{Heads of Transformer-Encoder}
	\label{fig:Heads_of_Transformer-Encoder}       
\end{figure}

Then we adjusted the number of heads in transformer-encoder. On the same datasets, the experiment results are shown in Figure~\ref{fig:Heads_of_Transformer-Encoder}. In fact, the number of heads from 1 to 4 has a less than 0.5(ms) effect on Test Time, and it is very slight. On DMSC\_V2 and Online\_shopping, the number of heads 2 can achieve the best accuracy and F1. On YF\_DianPing, the number of heads 1 and 2 are similar in accuracy and F1.


\begin{figure}
	\centering
	\subfigure[DMSC\_V2.]{
	\includegraphics[width=0.45\textwidth]{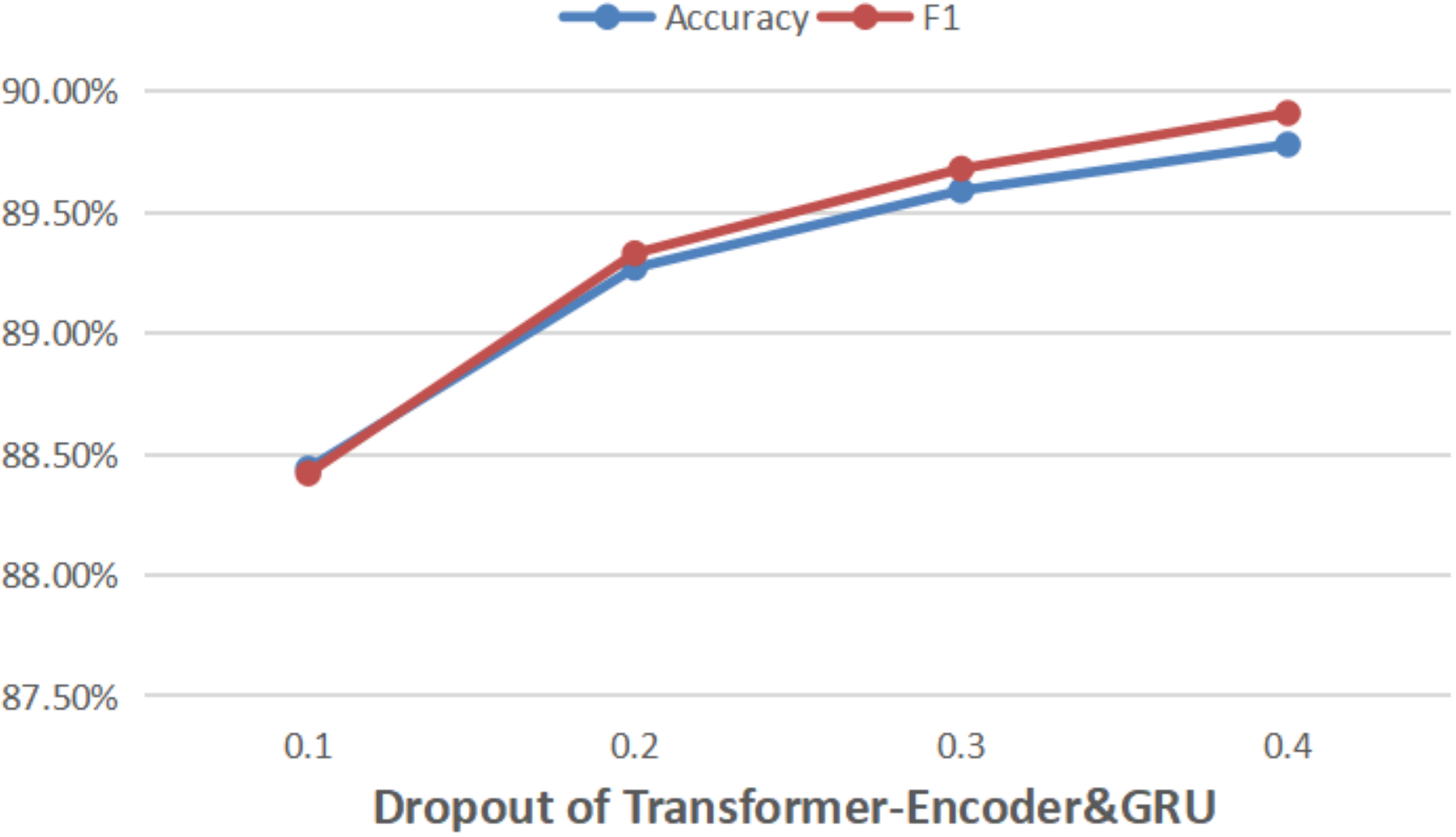}

    }
    \quad
	\subfigure[YF\_DianPing]{
	\includegraphics[width=0.45\textwidth]{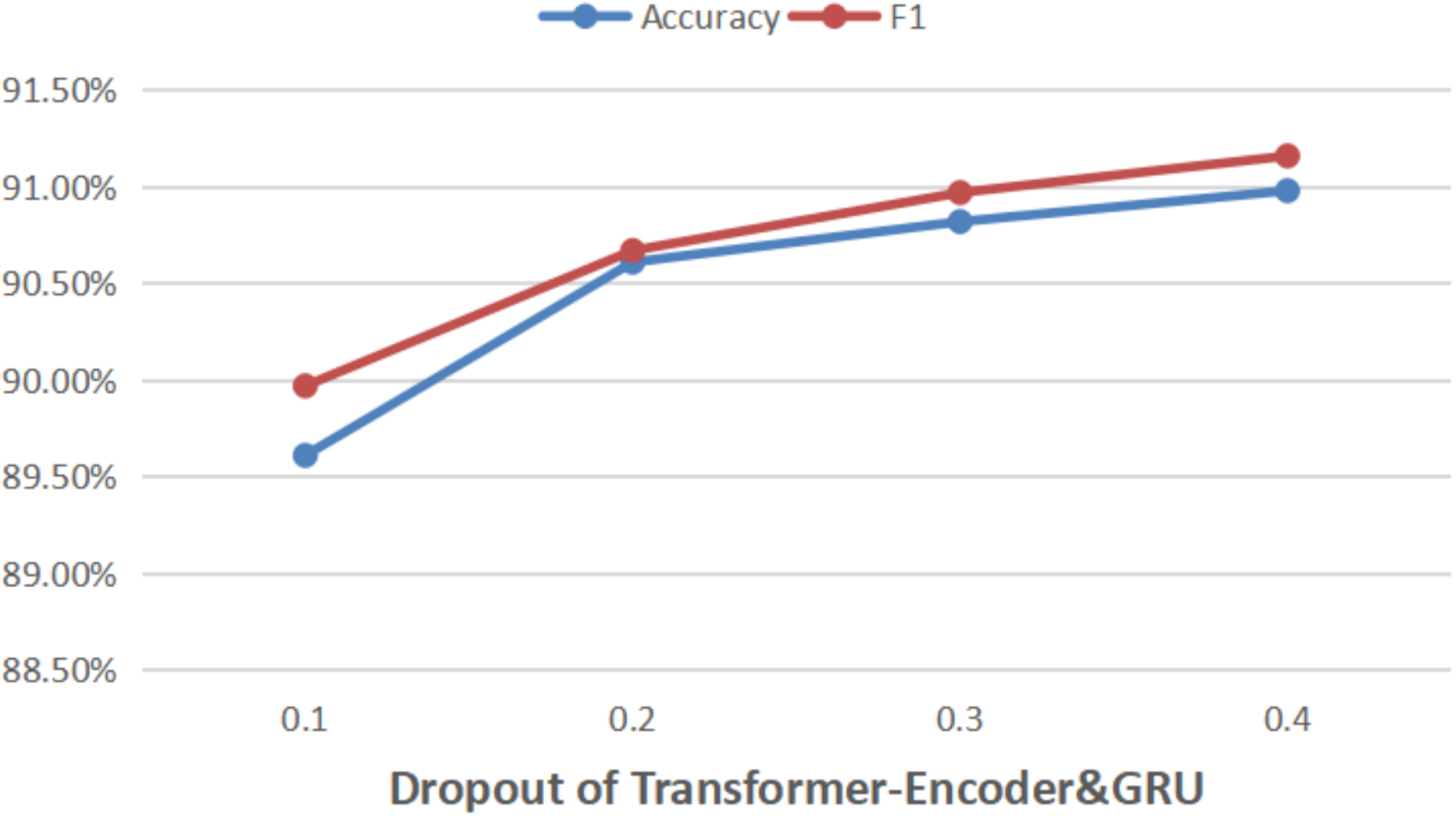}
    }
    \quad
	\subfigure[Online\_shopping\_10 cats Dataset]{
	\includegraphics[width=0.45\textwidth]{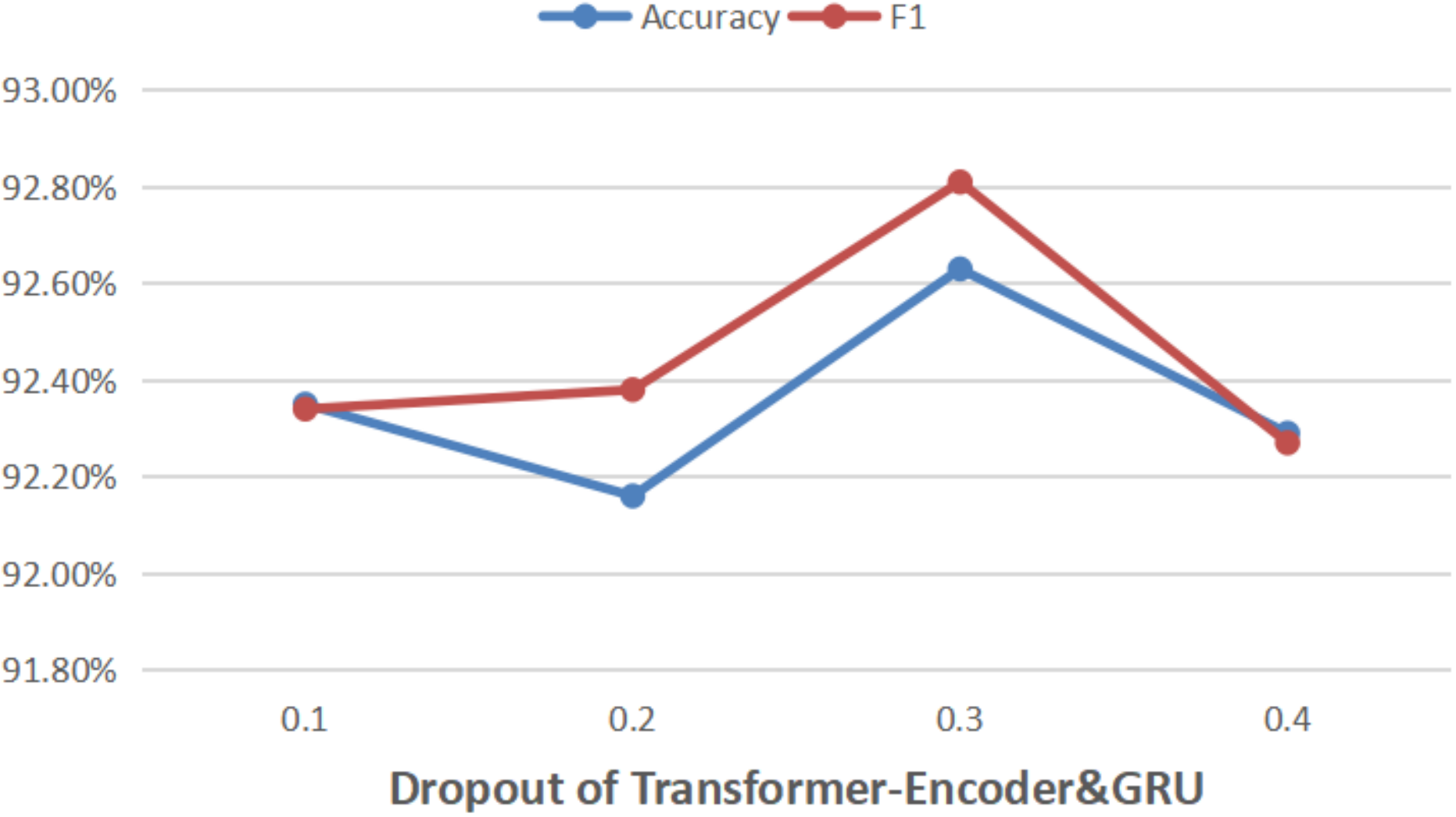}
    }
	\caption{Dropout of Transformer-Encoder \& GRU}
	\label{fig:Dropout_Transformer-Encoder}       
\end{figure}
Finally, we adjusted the dropout of transformer-encoder and GRU on the same datasets. Since dropout does not affect the complexity of the model, we removed Test Time, and the result is shown in Figure~\ref{fig:Dropout_Transformer-Encoder}. As the result shows, on DMSC\_V2 and YF\_DianPing, dropout of 0.4 is the best in accuracy and F1, while on Online\_shopping, 0.3 is the best.

In general, we can find:
\begin{enumerate}
    \item GRU is more suitable for combination with transformer-ecoder than LSTM and RNN.
    \item For T-E-GRU, adjusting the hidden size of transformer-encoder, heads of transformer-encoder and dropout has little effect on Test Time, while the accuracy and F1 are different depending on the datasets.
\end{enumerate}

\section{Conclusion}
On the three real Chinese review datasets which have not been strictly cleaned, we have retained punctuation with the ability to clauses as preprocessing. Then We compared our proposed T-E-GRU with various recurrent model and recurrent model with attention. On the test set, compared to the best performance in other models, T-E-GRU has 1\% improvement in accuracy and F1.

\bibliographystyle{plain}
\bibliography{ref}   

%

\end{document}